\documentclass[11pt,a4paper]{article}
\usepackage[hyperref]{emnlp-ijcnlp-2019}
\usepackage{times}
\usepackage{latexsym}
\usepackage{url}

\usepackage{graphicx}
\usepackage{epstopdf}
\usepackage{amsmath}
\usepackage{multirow}
\usepackage{amssymb}
 \usepackage{array}

\aclfinalcopy 

\title{Generative Dialogue Policy for Task-oriented Dialogue Systems}

\author{First Author \\
  Affiliation / Address line 1 \\
  Affiliation / Address line 2 \\
  Affiliation / Address line 3 \\
  {\tt email@domain} \\\And
  Second Author \\
  Affiliation / Address line 1 \\
  Affiliation / Address line 2 \\
  Affiliation / Address line 3 \\
  {\tt email@domain} \\}

\date{}

\begin{document}
\maketitle
\begin{abstract}
  There is an increasing demand for task-oriented dialogue systems which can assist users in various activities such as booking tickets and restaurant reservations.
  In order to complete dialogues effectively, dialogue policy plays a key role in task-oriented dialogue systems.\@
  As far as we know, the existing task-oriented dialogue systems obtain the dialogue policy through classification, which can assign either a dialogue act and its corresponding parameters or multiple dialogue acts without their corresponding parameters for a dialogue action\footnote{In the dialogue scenario, a dialogue action is generated by the learned dialogue policy in every turn. A dialogue act is the act label in task, such as \textit{offer} and \textit{request}. The parameter of a dialogue act is a collection of \textit{(slot=value)} pairs. In previous works, a dialogue action consists of one dialogue act and its parameters or multiple dialogue acts without their parameters. In this work, a dialogue action consists of multiple dialogue acts and their corresponding parameters. }.\@
  In fact, a good dialogue policy should construct multiple dialogue acts and their corresponding parameters at the same time.\@ However, it's hard for existing classification-based methods to achieve this goal.\@
  Thus, to address the issue above, we propose a novel generative dialogue policy learning method.\@ Specifically, the proposed method uses attention mechanism to find relevant segments of given dialogue context and input utterance, and then constructs the dialogue policy by a seq2seq way for task-oriented dialogue systems.\@
  Extensive experiments on two benchmark datasets show that the proposed model significantly outperforms the state-of-the-art baselines. 
  In addition, we have publicly released our codes\footnote{Github address is xxx (anonymous)}.
\end{abstract}

\section{Introduction}
  Task-oriented dialogue system is an important tool to build personal virtual assistants, which can help users to complete most of the daily tasks by interacting with devices via natural language. It's attracting increasing attention of researchers, and lots of works have been proposed in this area \cite{peng2018deep, eric2017copy, lipton2018bbq, young2013pomdp, wen2016network, lei2018sequicity, schatzmann2007statistical, schatzmann2007agenda}.
  
  The existing task-oriented dialogue systems usually consist of four components: (1) natural language understanding (NLU), it tries to identify the intent of a user;  
  (2) dialogue state tracker (DST), it keeps the track of user goals and constraints in every turn; (3) dialogue policy maker (DP), 
  it aims to generate the next available dialogue action;
  and (4) natural language generator (NLG), it generates a natural language response based on the dialogue action.  
  Among the four components, dialogue policy maker plays a key role in order to complete dialogues effectively, because it decides the next dialogue action to be executed. 
  
  As far as we know, the dialogue policy makers in most existing task-oriented dialogue systems just use the classifiers of the predefined acts to obtain dialogue policy \cite{peng2018deep, lipton2018bbq, wen2016network, liu2017end, liu2017iterative}.
  The classification-based dialogue policy learning methods can assign either only a dialogue act and its corresponding parameters \cite{su2016continuously, lipton2018bbq, peng2018deep} or multiple dialogue acts without their corresponding parameters for a dialogue action \cite{chi2017speaker}. However, all these existing methods cannot obtain multiple dialogue acts and their corresponding parameters for a dialogue action at the same time. 
  
  \begin{figure}[t]        
  \center{\includegraphics[width=7.6cm, height=4.5cm]  {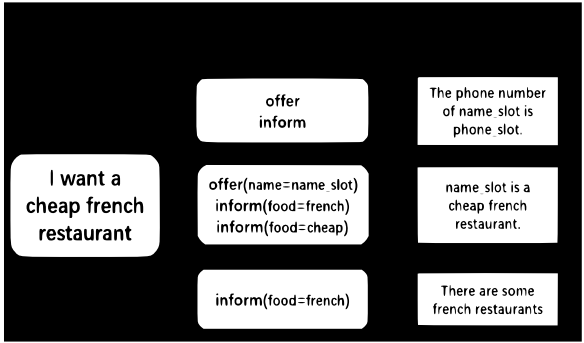}}        \caption{The examples in DSTC2 dataset, our proposed model can hold more information about dialogue policy than the classification models mentioned above. ``\textbf{MA, w/o P}" is the model that chooses multiple acts without corresponding parameters during dialogue police modeling, ``\textbf{w/o MA, P}" is the model that chooses only one act and its parameters.}
  \label{img:1}
  \end{figure}
  
  Intuitively, it will be more reasonable to construct multiple dialogue acts and their corresponding parameters for a dialogue action at the same time. For example, it can be shown that there are 49.4\% of turns in the DSTC2 dataset and 61.5\% of turns in the Maluuba dataset have multiple dialogue acts and their corresponding parameters as the dialogue action.  
  If multiple dialogue acts and their corresponding parameters can be obtained at the same time, the final response of task-oriented dialogue systems will become more accurate and effective. 
  For example, as shown in Figure \ref{img:1}, a user wants to get the name of a \textit{cheap french} restaurant.\@ The correct dialogue policy should generate three acts in current dialogue turn: \textit{offer(name=name\_slot)}, \textit{inform(food=french)} and \textit{inform(food=cheap)}. Thus, the user's real thought may be: ``\textit{name\_slot is a cheap french restaurant}". If losing the act \textit{offer}, the system may generate a response like ``\textit{There are some french restaurants}", which will be far from the user's goal.
  
  To address this challenge, we propose a Generative Dialogue Policy model (GDP) by casting the dialogue policy learning problem as a sequence optimization problem. 
  The proposed model generates a series of acts and their corresponding parameters by the learned dialogue policy.
  Specifically, our proposed model uses a recurrent neural network (RNN) as action decoder to construct dialogue policy maker instead of traditional classifiers. Attention mechanism is used to help the decoder decode dialogue acts and their corresponding parameters, and then the template-based natural language generator uses the results of the dialogue policy maker to choose an appropriate sentence template as the final response to the user. 
  
  Extensive experiments conducted on two benchmark datasets verify the effectiveness of our proposed method. Our contributions in this work are three-fold.
  \begin{itemize}
      \item The existing methods cannot construct multiple dialogue acts and their corresponding parameters at the same time. In this paper, We propose a novel generative dialogue policy model to solve the problem.
      \item The extensive experiments demonstrate that the proposed model significantly outperforms the state-of-the-art baselines on two benchmarks.
      \item We publicly release the source code.
  \end{itemize}
  
\section{Related Work}

Usually, the existing task-oriented dialogue systems use a pipeline of four separate modules: natural language understanding, dialogue belief tracker, dialogue policy and natural language generator.
Among these four modules, dialogue policy maker plays a key role in task-oriented dialogue systems, which generates the next dialogue action.

As far as we know, nearly all the existing approaches obtain the dialogue policy by using the classifiers of all predefined dialogue acts \cite{su2017sample, jurvcivcek2011natural}. There are usually two kinds of dialogue policy learning methods. 
One constructs a dialogue act and its corresponding parameters for a dialogue action. For example, \citet{peng2018deep} constructs a simple classifier for all the predefined dialogue acts. \citet{lipton2018bbq} build a complex classifier for some predefined dialogue acts, addtionally \citet{lipton2018bbq} adds two acts for each parameter: one to inform its value and the other to request it.\@ The other obtains the dialogue policy by using multi-label classification to consider multiple dialogue acts without their parameters. \citet{chi2017speaker} performs multi-label multi-class classification for dialogue policy learning and then the multiple acts can be decided based on a threshold.
Based on these classifiers, the reinforcement learning can be used to further update the dialogue policy of task-oriented dialogue systems \cite{young2013pomdp, cuayahuitl2015strategic, liu2017iterative}.

In the real scene, an correct dialogue action usually consists of multiple dialogue acts and their corresponding parameters. However, it is very hard for existing classification-based dialogue policy maker to achieve this goal.
Thus, in this paper we propose a novel generative dialogue policy maker to address this issue by casting the dialogue policy learning problem as a sequence optimization problem.

\section{Technical Background}
\subsection{Encoder-Decoder Seq2Seq Models}
Seq2Seq model was first introduced by \citet{cho2014learning} for statistical machine translation. It uses two recurrent neural networks (RNN) to solve the sequence-to-sequence mapping problem. One called encoder encodes the user utterance into a dense vector representing its semantics, the other called decoder decodes this vector to the target sentence. 
Now Seq2Seq framework has already been used in task-oriented dialog systems such as \cite{wen2016network} and \cite{eric2017copy}, and shows the challenging performance. In the Seq2Seq model, given the user utterance $Q=(x_1, x_2, ..., x_n)$, the encoder squeezes it into a context vector $C$ and then used by decoder to generate the response $R=(y_1, y_2, ..., y_m)$ word by word by maximizing the generation probability of $R$ conditioned on $Q$. The objective function of Seq2Seq can be written as:

\begin{equation}\label{1}
\begin{split}
& p(y_1,...,y_m|x_1,...,x_n) = \\
& p(y_1|C) \prod_{t=2}^T p(y_t|C,y_1,...,y_{t-1}) \\
\end{split}
\end{equation} In particular, the encoder RNN produces the context vector $C$ by doing calculation below:

\begin{equation}\label{2}
\begin{split}
& h_t=f(x_t,h_{t-1}) \\
& C=h_n \\
\end{split}
\end{equation} The $h_t$ is the hidden state of the encoder RNN at time step $t$ and $f$ is the non-linear transformation which can be a long-short term memory unit LSTM \cite{hochreiter1997long} or a gated recurrent unit GRU \cite{cho2014learning}. In this paper, we implement $f$ by using GRU.

The decoder RNN generates each word in reply conditioned on the context vector $C$. The probability distribution of candidate words at every time step $t$ is calculated as:

\begin{equation}\label{3}
\begin{split}
& s_t=f(y_{t-1}, s_{t-1}, C) \\
& y_t={\rm softmax}(s_t, y_{t-1}) \\
\end{split}
\end{equation} The $s_t$ is the hidden state of decoder RNN at time step $t$ and $y_{t-1}$ is the generated word in the reply at time $t-1$ calculated by softmax operations.

\subsection{Attention Mechanism}
Attention mechanisms \cite{bahdanau2014neural} have been proved to improved effectively the generation quality for the Seq2Seq framework. In Seq2Seq with attention, each $y_i$ corresponds to a context vector $C_i$ which is calculated dynamically. It is a weighted average of all hidden states of the encoder RNN. Formally, $C_i$ is defined as $C_i=\sum_{j=1}^{n} \alpha_{ij}h_j$, where $\alpha_{ij}$ is given by:

\begin{equation} \label{4}
\begin{split}
& \alpha_{ij}=\frac{{\rm exp}(e_{ij})}{\sum_{k=1}^{n}{\rm exp}(e_{ik})} \\
& e_{ij}=\eta(s_{i-1},h_j) \\
\end{split}
\end{equation} where $s_{i-1}$ is the last hidden state of the decoder, the $\eta$ is often implemented as a multi-layer-perceptron (MLP) with tanh as the activation function.

\section{Generative Dialogue Policy}
\begin{figure*}[t]        
 \center{\includegraphics[width=13cm]  {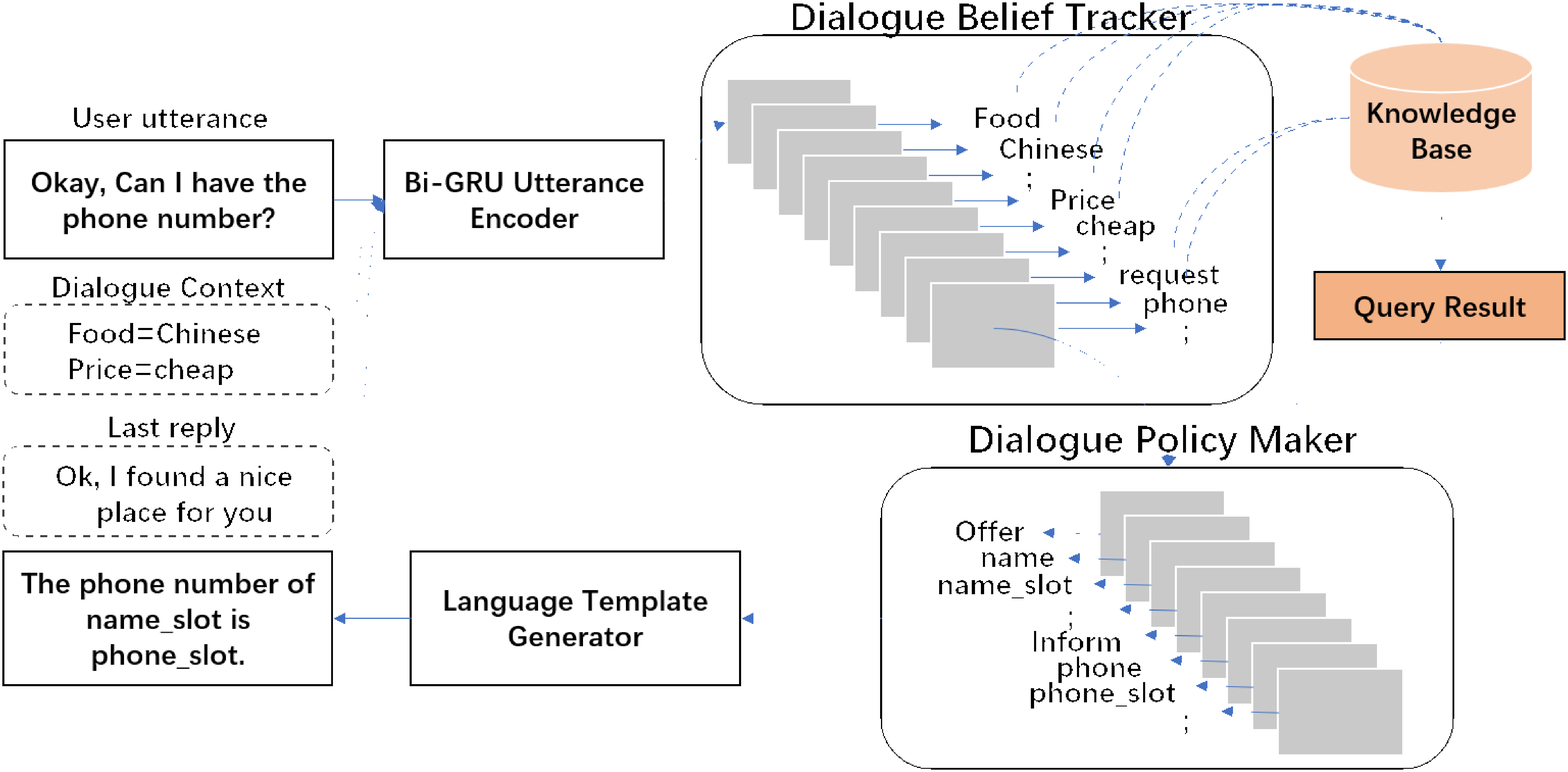}}        
 \caption{GDP overview. The utterance encoder encodes the user utterance, the dialogue context and the last reply of the systems into the dense vector. As for dialogue belief tracker, we use the approach of \citet{lei2018sequicity} to generate dialogue context. Then this information will be used to search the knowledge base. Based on the user's intents and query results, dialogue policy maker generates the next dialogue action by using our RNN-based proposed method.}
 \label{img:2}
 \end{figure*}

Figure \ref{img:2} shows the overall system architecture of the proposed GDP model. Our model contains five main components: (1) utterance encoder; (2) dialogue belief tracker; (3) dialogue policy maker; (4) knowledge base; (5) template-based natural language generator. Next, we will describe each component of our proposed GDP model in detail.

\subsection{Notations and Task Formulation}
Given the user utterance $U_t$ at turn $t$ and the dialogue context $C_{t-1}$ which contains the result of the dialogue belief tracker at turn $t-1$, the task-oriented dialog system needs to generate user's intents $C_t$ by dialogue belief tracker and then uses this information to get the knowledge base query result $k_t \in \mathbb{R}^k$. Then the model needs to generate the next dialogue action $A_t$ based on $k_t$, $U_t$ and $C_t$. The natural language generator provides the template-based response $R_t$ as the final reply by using $A_t$. The $U_t$ and $C_t$ are the sequences, $k_t$ is a one-hot vector representing the number of the query results. For baselines, in this paper, the $A_t$ is the classification result of the next dialogue action, but in our proposed model it's a sequence which contains multiple acts and their corresponding parameters.

\subsection{Utterance Encoder}
A bidirectional GRU is used to encode the user utterance  $U_t$, the last turn response $R_{t-1}$ made by the system and the dialogue context $C_{t-1}$ into a continuous representation. The vector is generated by concatenating the last forward and backward GRU states. $U_t = (w_1, w_2, ..., w_{T_m})$ is the user utterance at turn $t$. $C_{t-1}=(c_1, c_2, ..., c_{T_n})$ is the dialogue context made by dialogue belief tracker at $t-1$ turn. $R_{t-1}$ is the response made by our task-oriented dialogue system at last turn. Then the words of $[C_{t-1}, R_{t-1}, U_t]$ are firstly mapped into an embedding space and further serve as the inputs of each step to the bidirectional GRU. Let $n$ denotes the number of words in the sequence $[C_{t-1}, R_{t-1}, U_t]$. 
The $\overrightarrow{h_{t'}^u}$ and $\overleftarrow{h_{t'}^u}$ represent the forward and backward GRU state outputs at time step $t'$. 
The encoder output of timestep $i$ denote as $\overline{h_i^u}$.

\begin{equation} \label{5}
\begin{split}
& H_u = {\rm BiGRU}(e([C_{t-1}, R_{t-1}, U_t])) \\
& \overline{h_i^u} = \overrightarrow{h_i^u} + \overleftarrow{h_i^u}, \overline{h_i^u} \in \mathbb{R}^{d_h}\\
& H_u = \{\overline{h_1^u}, \overline{h_2^u}, ..., \overline{h_n^u}\} \\
\end{split}
\end{equation} where $e([C_{t-1}, R_{t-1}, U_t])$ is the embedding of the input sequence, $d_h$ is the hidden size of the GRU. $H_u$ contains the encoder hidden state of each timestep, which will be used by attention mechanism in dialogue policy maker.

\subsection{Dialogue State Tracker}
Dialogue state tracker maintains the state of a conversation and collects the user's goals during the dialogue. Recent work successfully represents this component as discriminative classifiers. \citet{lei2018sequicity} verified that the generation is a better way to model the dialogue state tracker. 

Specifically, we use a GRU as the generator to decode the $C_t$ of current turn. In order to capture user intent information accurately, the basic attention mechanism is calculated when the decoder decodes the $C_t$ at each step, which is the same as the Eq. (\ref{4}).
\begin{equation} \label{5}
\begin{split}
& c_i = \sum_{j=1}^n \alpha_{ij}\overline{h_j^u} \\
& h_i^d = {\rm GRU}(h^d_{i-1}, e(y_{i-1}^d)), h_i^d \in \mathbb{R}^{d_h} \\
& y_i^d = {\rm softmax}([h_i^d, c_i]) \\
& H_d = \{h_1^d, h_2^d, ..., h_m^d\} \\
& C_t = \{y_1^d, y_2^d, ..., y_m^d\} \\
\end{split}
\end{equation} where $m$ is the length of $C_t$, $e(y_i)$ is the embedding of the token, $d_h$ is the hidden size of the GRU and the hidden state at $i$ timestep of the RNN in dialogue state tracker denote as $h_i^d$. The decoded token at step $i$ denotes as $y_i^d$.

\subsection{Knowledge Base}
Knowledge base is a database that stores information about the related task. For example, in the restaurant reservation, a knowledge base stores the information of all the restaurants, such as location and price. After dialogue belief tracker, the $C_t$ will be used as the constraints to search the results in knowledge base. Then the one-hot vector $k_t$ will be produced when the system gets the number of the results.

The search result $k_t$ has a great influence on dialogue policy. For example, if the result has multiple matches, the system should request more constraints of the user. In practice, let $k_t$ be an one-hot vector of 20 dimensions to represent the number of query results. Then $k_t$ will be used as the cue for dialogue policy maker. 

\subsection{Dialogue Policy Maker}
In task-oriented dialogue systems, supervised classification is a straightforward solution for dialogue policy modeling. However, we observe that classification cannot hold enough information for dialogue policy modeling. The generative approach is another way to model the dialogue policy maker for task-oriented dialogue systems, which generates the next dialogue acts and their corresponding parameters based on the dialogue context word by word. Thus the generative approach converts the dialogue policy learning problem into a sequence optimization problem.

The dialogue policy maker generates the next dialogue action $A_t$ based on $k_t$ and $[H_u, H_d]$. Our proposed model uses the GRU as the action decoder to decode the acts and their parameters for the response. Particularly, at step $i$, for decoding $y_i^p$ of $A_t$, the decoder GRU takes the embedding of $y_{i-1}^p$ to generate a hidden vector $h_i^p$. Basic attention mechanism is calculated.

\begin{equation} \label{6}
\begin{split}
& c_u = \sum_{j=1}^{n} \alpha_{ij}h_j^u; c_d = \sum_{j=1}^{m} \alpha_{ij}h_j^d\\
& h_i^p = {\rm GRU}(h_{i-1}^p, e(y_{i-1})) \\
\end{split}
\end{equation} where $e$ is the embedding of the token, $c_u$ is the context vector of the input utterance and $c_d$ is the context vector of the dialogue state tracker. $h_i^p$ is the hidden state of the GRU in dialogue policy maker at $i$ timestep.

\begin{equation} \label{7}
\begin{split}
& y_i^p = {\rm softmax}(O [h_i^p, k_t, c_u, c_d]) \\
& A_t = \{y_1^p, y_2^p, ..., y_k^p\} \\
\end{split}
\end{equation} where $y_i^p$ is the token decoded at $i$ timestep. And the final results of dialogue policy maker denote as $A_t$, and the $k$ is the length of it.
In our proposed model, the dialogue policy maker can be viewed as a decoder of the seq2seq model conditioned on $[C_{t-1},R_{t-1},U_t]$ and $k_t$. 

\subsection{Nature Language Generator}
After getting the dialogue action $A_t$ by the learned dialogue policy maker, the task-oriented dialogue system needs to generate an appropriate response $R_t$ for users. 
We construct the natural language generator by using template sentences. For each dataset, we extract all the system responses, then we manually modify responses to construct the sentence templates for task-oriented dialogue systems.
In our proposed model, the sequence of the acts and parameters $A_t$ will be used for searching appropriate template. However, the classification-based baselines use the categories of acts and their corresponding parameters to search the corresponding template.

\subsection{Training}
In supervised learning, because our proposed model is built in a seq2seq way, the standard cross entropy is adopted as our objective function to train dialogue belief tracker and dialogue policy maker.

\begin{equation} \label{9}
J=\sum_{j=1}^m y_j^d log P_j(y_j^d) + \sum_{j=1}^k y_j^p log P_j(y_j^p)
\end{equation}

After supervised learning, the dialogue policy can be further updated by using reinforcement learning. 
In the context of reinforcement learning, the  decoder of dialogue policy maker can be viewed as a policy network, denoted as $\pi_{\theta}(y_j)$ for decoding $y_j$, $\theta$ is the parameters of the decoder. Accordingly, the hidden state created by GRU is the corresponding state, and the choice of the current token $y_j$ is an action\footnote{The action here is different from the dialogue action. It's a concept of the reinforcement learning.}. 

Reward function is also very important for reinforcement learning when decoding every token. To encourage our policy maker to generate correct acts and their corresponding parameters, we set the reward function as follows: once the dialogue acts and their parameters are decoded correctly, the reward is 2; otherwise, the reward is -5; only the label of the dialogue act is decoded correctly but parameters is wrong, the reward is 1; $\lambda$ is a decay parameter. More details are shown in Sec \ref{ref:1}. In our proposed model, rewards can only be obtained at the end of decoding $A_t$. In order to get the rewards at each decoding step, we sample some results $A_t$ after choosing $y_j$, and the reward of $y_j$ is set as the average of all the sampled results' rewards. 

In order to ensure that the model's performance is stable during the fine-tuning phase of reinforcement learning, we freeze the parameters of user utterance and dialogue belief tracker, only the parameters of the dialogue policy maker will be optimized by reinforcement learning. Policy gradient algorithm REINFORCE \cite{williams1992simple} is used for pretrained dialogue policy maker:

\begin{equation} \label{10}
J=-\frac{1}{m}\sum_{j=1}^m r(y_j) \frac{\partial \log \pi_{\theta}(y_j)}{\partial \theta}
\end{equation} where the $m$ is the length of the decoded action. The objective function $J$ can be optimized by gradient descent.

\section{Experiments}
We evaluate the performance of the proposed model in three aspects: (1) the accuracy of the dialogue state tracker, it aims to show the impact of the dialogue state tracker on the dialogue policy maker; (2) the accuracy of dialogue policy maker, it aims to explain the performance of different methods of constructing dialogue policy; (3) the quality of the final response, it aims to explain the impact of the dialogue policy on the final dialogue response.\@
The evaluation metrics are listed as follows:

\begin{itemize}
    \item \textbf{BPRA}: Belief Per-Response Accuracy (BPRA) tests the ability to generate the correct user intents during the dialogue. This metric is used to evaluate the accuracy of dialogue belief tracker \cite{eric2017copy}.
    \item \textbf{APRA}: Action Per-Response Accuracy (APRA) evaluates the per-turn accuracy of the dialogue actions generated by dialogue policy maker. 
    For baselines, \textbf{APRA} evaluates the classification accuracy of the dialogue policy maker. But our model actually generates each individual token of actions, and we consider a prediction to be correct only if every token of the model output matches the corresponding token in the ground truth.
    \item \textbf{BLEU} \cite{papineni2002bleu}: The metric evaluates the quality of the final response generated by natural language generator. The metric is usually used to measure the performance of the task-oriented dialogue system.
\end{itemize}

We also choose the following metrics to evaluate the efficiency of training the model:

\begin{itemize}
    \item \textbf{$\mathbf{Time_{full}}$}: The time for training the whole model, which is important for industry settings.
    \item \textbf{$\mathbf{Time_{DP}}$}: The time for training the dialogue policy maker in a task-oriented dialogue system.
\end{itemize}

\subsection{Datasets}
\begin{table}[]
    \begin{tabular}{|p{2.3cm}|p{4.7cm}|}
    \hline
    \textbf{Dataset} & DSTC2 \\
    \hline
    \textbf{Size} & Train:1612,Test:506,Dev:1117\\ 
    \hline
    \textbf{Domains} & restaurant reservation\\
    \hline
    \textbf{Actions} & 11. offer, inform, request etc.\\
    \hline
    \textbf{Slots} & 8. area, food, price etc.\\
    \hline
    \textbf{Distinct value} & 212\\
    \hline
    \end{tabular}
    
   \begin{tabular}{|p{2.3cm}|p{4.7cm}|}
    \hline
    \textbf{Dataset} & Maluuba \\
    \hline
    \textbf{Size} & Train:8621,Test:478,Dev:480\\ 
    \hline
    \textbf{Domains} & travel booking\\
    \hline
    \textbf{Actions} & 16. offer, inform, request etc.\\
    \hline
    \textbf{Slots} & 60. startdate, enddate etc.\\
    \hline
    \textbf{Distinct value} & inf (continuous values)\\
    \hline
    \end{tabular}
    \caption{The details of DSTC2 and Maluuba dataset. The Maluuba dataset is more complex than DSTC2, and has some continuous value space such as time and price which is hard to solve for classification model.}
    \label{tab:3}
    \end{table}
    
We adopt the DSTC2 \cite{henderson2014second} dataset and Maluuba \cite{asri2017frames} dataset to evaluate our proposed model. Both of them are the benchmark datasets for building the task-oriented dialog systems. Specifically, the DSTC2 is a human-machine dataset in the single domain of restaurant searching. The Maluuba is a very complex human-human dataset in travel booking domain which contains more slots and values than DSTC2. Detailed slot information in each dataset is shown in Table \ref{tab:3}.

\begin{table*}[h]
\small
\begin{center}
\begin{tabular}{|c||p{0.8cm}|p{0.8cm}<{\centering}|p{0.8cm}|p{1.1cm}<{\centering}|p{1.1cm}<{\centering}||p{0.8cm}|p{0.8cm}<{\centering}|p{0.8cm}|p{1.1cm}<{\centering}|p{1.1cm}<{\centering}|}
\hline
\multirow{2}{*}{\textbf{Models}} & \multicolumn{5}{c||}{DSTC2} & \multicolumn{5}{c|}{Maluuba} \\ \cline{2-11} 
    & BPRA & APRA  & BLEU & $Time_{full}$ & $Time_{DP}$ & BPRA  & APRA & BLEU  & $Time_{full}$ & $Time_{DP}$ \\ \hline
\textbf{E2ECM}                   & 0.9689  & -       & 0.1782 & \textbf{42.30 m} & \textbf{0.78 m} & 0.7458   & -       & 0.0797 & \textbf{45.81 m} & \textbf{0.84 m}  \\ \hline
\textbf{CDM}                     & 0.9704  & 0.2791  & 0.2039 & 45.71 m & 2.96 m & 0.6771   & 0.1542  & 0.0704 & 50.22 m & 3.25 m \\ \hline
\textbf{GDP}                     & \textbf{0.9719}  & \textbf{0.5732}  & \textbf{0.2847} & 46.43 m & 9.63 m & \textbf{0.7500}   & \textbf{0.4512}  & \textbf{0.1156} & 55.51 m & 11.49 m  \\ \hline
\hline
\textbf{E2ECM+RL}                & 0.9689  & -       & 0.1823 & \textbf{30.01 m} & \textbf{30.01 m} & 0.7458   & -       & 0.0799 & 35.13 m & 35.13 m \\ \hline
\textbf{CDM+RL}                  & 0.9704  & 0.2873  & 0.2088 & 101.0 m & 101.0 m & 0.6771   & 0.1625  & 0.0734 & \textbf{29.00 m} & \textbf{29.00 m} \\ \hline
\textbf{GDP+RL}
& \textbf{0.9719}  & \textbf{0.5766}  & \textbf{0.2879} & 98.07 m & 98.07 m & \textbf{0.7500}   & \textbf{0.4521}  & \textbf{0.1226} & 134.8 m & 134.8 m  \\ \hline

\end{tabular}
\end{center}
\caption{The performance of baselines and proposed model on DSTC2 and Maluuba dataset. $Time_{full}$ is the time spent on training the whole model, $Time_{DP}$ is the time spent on training the dialogue policy maker.}
\label{tab:4}
\end{table*}

\subsection{Baselines}
For comparison, we choose two state-of-the-art baselines and their variants.

\begin{itemize}
    \item \textbf{E2ECM} \cite{chi2017speaker}: In dialogue policy maker, it adopts a classic classification for skeletal sentence template. In our implement, we construct multiple binary classifications for each act to search the sentence template according to the work proposed by  \citet{chi2017speaker}.
    \item \textbf{CDM} \cite{su2016continuously}: This approach designs a group of classifications (two multi-class classifications and some binary classifications) to model the dialogue policy.
    \item \textbf{E2ECM+RL}: It fine tunes the classification parameters of the dialogue policy by REINFORCE \cite{williams1992simple}.
    \item \textbf{CDM+RL}: It fine tunes the classification of the act and corresponding parameters by REINFORCE \cite{williams1992simple}.
\end{itemize}

In order to verify the performance of the dialogue policy maker, the utterance encoder and dialogue belief tracker of our proposed model and baselines is the same, only dialogue policy maker is different.

\subsection{Parameters settings} \label{ref:1}

For all models, the hidden size of dialogue belief tracker and utterance encoder is 350, and the embedding size $d_{emb}$ is set to 300. For our proposed model, the hidden size of decoder in dialogue policy maker is 150. The vocabulary size $|V|$ is 540 for DSTC2 and 4712 for Maluuba. And the size of $k_t$ is set to 20. An Adam optimizer \cite{kingma2014adam} is used for training our models and baselines, with a learning rate of 0.001 for supervised training and 0.0001 for reinforcement learning. In reinforcement learning, the decay parameter $\lambda$ is set to 0.8. The weight decay is set to 0.001. And early stopping is performed on developing set.

\subsection{Experimental Results}
The experimental results of the proposed model and baselines will be analyzed from the following aspects.

\textbf{BPRA Results:} As shown in Table \ref{tab:4}, most of the models have similar performance on BPRA on these two datasets, which can guarantee a consistent impact on the dialogue policy maker. 
All the models perform very well in BPRA on DSTC2 dataset. On Maluuba dataset, the BPRA decreases because of the complex domains. We can notice that BPRA of CDM is slightly worse than other models on Maluuba dataset, the reason is that the CDM's dialogue policy maker contains lots of classifications and has the bigger loss than other models because of complex domains, which affects the training of the dialogue belief tracker.

\textbf{APRA Results:} Compared with baselines, GDP achieves the best performance in APRA on two datasets.  
It can be noted that we do not compare with the E2ECM baseline in APRA. 
E2ECM only uses a simple classifier to recognize the label of the acts and ignores the parameters information. 
In our experiment, APRA of E2ECM is slightly better than our method.
Considering the lack of parameters of the acts, it's unfair for our GDP method. 
Furthermore, the CDM baseline considers the parameters of the act. But GDP is far better than CDM in supervised learning and reinforcement learning.  

\begin{table*}[h]
\resizebox{\textwidth}{!}{%
\begin{tabular}{|p{5cm}<{\centering}|p{3.5cm}<{\centering}|p{3.5cm}<{\centering}|p{3cm}<{\centering}|p{3.5cm}<{\centering}|}
\hline
\multicolumn{1}{|c|}{\textbf{Dilogue Context}}  
& \textbf{Ground Truth} & \textbf{GDP} & \textbf{E2ECM} & \textbf{CDM} \\ \hline

\multirow{2}{*}{\begin{tabular}[c]{@{}p{5cm}<{\centering}@{}}\textbf{Inf}: cheap, east; \textbf{sys}: name\_slot is a nice place in the east of town and the price is cheap; \textbf{user}: what's the address?\end{tabular}} 
& \begin{tabular}[c]{@{}p{3.5cm}<{\centering}@{}}offer name name\_slot\\ inform addr addr\_slot\end{tabular} 
& \begin{tabular}[c]{@{}p{3.5cm}<{\centering}@{}}offer name name\_slot\\ inform addr addr\_slot\end{tabular} 
& \begin{tabular}[c]{@{}p{3cm}<{\centering}@{}}inform\\ offer\end{tabular}
& offer name name\_slot\\

\cline{2-5}

& \begin{tabular}[c]{@{}p{3.5cm}<{\centering}@{}}sure, name\_slot is \\on addr\_slot\end{tabular}
& \begin{tabular}[c]{@{}p{3.5cm}<{\centering}@{}}sure, name\_slot is \\on addr\_slot\end{tabular}
& \begin{tabular}[c]{@{}p{3cm}<{\centering}@{}}name\_slot is a nice place in the east of \\the town\end{tabular}
& \begin{tabular}[c]{@{}p{3.5cm}<{\centering}@{}}name\_slot is a \\nice place\end{tabular} \\ 
\hline
\end{tabular}%
}
\caption{Case Study on DSTC2 dataset. The first column is the \textbf{Dialogue Context} of this case, it contains three parts: (1) \textbf{Inf} is the user's intent captured by dialogue state tracker; (2)  \textbf{sys} is the system response at last turn; (3) \textbf{user} is the user utterance in this turn. The second column to the fifth column has two rows, above is the action made by the learned dialogue policy maker below is the final response made by template-based generator.}
\label{tab:6}
\end{table*}

\textbf{BLEU Results:} GDP significantly outperforms the baselines on BLEU. 
As mentioned above, E2ECM is actually slightly better than GDP in APRA. But in fact, we can find that the language quality of the response generated by GDP is still better than E2ECM, which proves that lack of enough parameters information makes it difficult to find the appropriate sentence template in NLG. 
It can be found that the BLEU of all models is very poor on Maluuba dataset. The reason is that Maluuba is a human-human task-oriented dialogue dataset, the utterances are very flexible, the natural language generator for all methods is difficult to generate an accurate utterance based on the context. And DSTC2 is a human-machine dialog dataset. The response is very regular so the effectiveness of NLG will be better than that of Maluuba. But from the results, the GDP is still better than the baselines on Maluuba dataset, which also verifies that our proposed method is more accurate in modeling dialogue policy on complex domains than the classification-based methods.

\begin{figure}[h]        
 \includegraphics[width=0.5\textwidth]{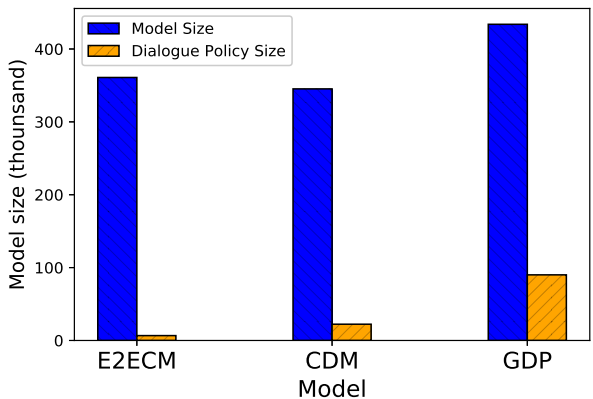}      
 \caption{The number of the parameters. GDP has the bigger model size and more dialogue policy parameters because of the RNN-based dialogue policy maker.}
 \label{img:3}
\end{figure}

\textbf{Time and Model Size:} 
In order to obtain more accurate and complete dialogue policy for task-oriented dialogue systems, the proposed model has more parameters on the dialogue policy maker than baselines.
As shown in Figure \ref{img:3}, E2ECM has the minimal dialogue policy parameters because of the simple classification. It needs minimum training time, but the performance of E2ECM is bad. 
The number of parameters in the CDM model is slightly larger than E2ECM. However, because both of them are classification methods, they all lose some important information about dialogue policy. Therefore, we can see from the experimental results that the quality of CDM's dialogue policy is as bad as E2ECM.
The number of dialogue policy maker's parameters in GDP model is much larger than baselines. Although the proposed model need more time to be optimized by supervised learning and reinforcement learning, the performance is much better than all baselines.


\subsection{Case Study}

Table \ref{tab:6} illustrates an example of our proposed model and baselines on DSTC2 dataset. In this example, a user's goal is to find a \textit{cheap restaurant in the east part of the town}. In the current turn, the user wants to get the address of the restaurant.

E2ECM chooses the \textit{inform} and \textit{offer} acts accurately, but the lack of the \textit{inform}'s parameters makes the final output deviate from the user's goal. CDM generates the parameters of \textit{offer} successfully, but the lack of the information of \textit{inform} also leads to a bad result. By contrast, the proposed model GDP can generate all the acts and their corresponding parameters as the dialogue action. Interestingly, the final result of GDP is exactly the same as the ground truth, which verifies that the proposed model is better than the state-of-the-art baselines.

\section{Conclusion}
In this paper, we propose a novel model named GDP. Our proposed model treats the dialogue policy modeling as the generative task instead of the discriminative task which can hold more information for dialogue policy modeling. We evaluate the GDP on two benchmark task-oriented dialogue datasets. Extensive experiments show that GDP outperforms the existing classification-based methods on both action accuracy and BLEU. 


\bibliography{emnlp-ijcnlp-2019}
\bibliographystyle{acl_natbib}

\appendix

\end{document}